\documentclass[letterpaper, 10 pt, conference]{ieeeconf}
\IEEEoverridecommandlockouts    % This command is only needed if
\usepackage{graphics}           
\usepackage{times}              
\usepackage{amsmath}            
\usepackage{amssymb}            
\usepackage{graphicx}
\graphicspath{{pics/}}
\usepackage{algorithm}
\usepackage[noend]{algpseudocode}
\usepackage{booktabs}
\usepackage{color}
\usepackage{listings}
\usepackage{subfiles}
\usepackage{hyperref}

\usepackage{siunitx}
\usepackage[nolist]{acronym}
\usepackage{multirow, multicol}
\usepackage{transparent}

\title{\LARGE \bf Transformer-based Monte Carlo Localization in Construction Meshes}

\author{Linus Kramer \and William Talbot \and Olga Vysotska \and Marco Hutter % <-this % stops a space
  \thanks{All authors are with the Robotic Systems Laboratory, ETH Zurich, Zurich 8092, Switzerland (e-mail: kramerl@ethz.ch; wtalbot@ethz.ch; ovysotska@ethz.ch; mahutter@ethz.ch)
  }%
}

\begin{document}
\thispagestyle{empty}
\pagestyle{empty}
\maketitle

\begin{acronym}
\acro{bim}[BIM]{Building Information Modeling}
\acro{mcl}[MCL]{Monte Carlo Localization}
\acro{icp}[ICP]{Iterative Closest Point}
\acro{mog}[MoG]{Mixture of Gaussians}
\end{acronym}

%%%%%%%%%%%%%%%%%%%%%%%%%%%%%%%%%%%%%%%%%%%%%%%%%%%%%%%%%%%%%%%%%%%%%%%%%%%%%%%%
\begin{abstract}

To be able to perform inspection or digitization tasks, mobile robots on construction sites must be able to localize themselves reliably with respect to a global reference frame that is shared with a building map. Similar room layouts and low-texture surfaces pose a challenge for existing LiDAR- and vision-based localization methods. We approach this problem with a LiDAR-based global relocalization system that estimates the robot’s pose relative to a building mesh and combines a PointNet++ encoder with a place recognition decoder, whose outputs serve as a learned observation model within a \ac{mcl} framework. The pipeline is trained exclusively on synthetic LiDAR scans obtained by simulating the robot’s sensors inside the building mesh. 
Our approach is robust in ambiguous environments due to an uncertainty-aware decoder that scales positional likelihoods and a resampling strategy that injects model hypotheses into the particle set, enabling recovery from potential particle depletion. Evaluations on real-world datasets show that our method outperforms both diffusion-based and ScanContext++ baselines while maintaining fast inference (18 ms per call), demonstrating the practicality of synthetic-data training for mesh-referenced global localization in construction robotics.

\end{abstract}

%%%%%%%%%%%%%%%%%%%%%%%%%%%%%%%%%%%%%%%%%%%%%%%%%%%%%%%%%%%%%%%%%%%%%%%%%%%%%%%%
\section{Introduction}
\label{sec:intro}

Inspection and progress monitoring on construction sites still heavily rely on manual work and coordination between people and resources. Tasks such as progress tracking and reporting are often time-consuming and less accurate since they depend on subjective human judgment. Thus, automating inspection and monitoring has been identified as a key enabler for improving efficiency in the construction industry~\cite{samsami2024systematic, qureshi2023automated}.
We pursue this automation task using a quadrupedal robot because it can navigate unstructured and uneven terrain, climb stairs, and maintain stability in environments that are inaccessible to wheeled or tracked platforms~\cite{7758092}. These capabilities make it particularly well-suited for construction sites, where obstacles, debris, and incomplete infrastructure are common.

A crucial component for automating deployments of quadrupedal robots on construction sites is the ability to perform global localization in such complex environments. A robust global localization system enables the robot to start the inspection missions without a priori known initial location and quickly recover the position in the map in case the tracking system fails.

\begin{figure}[t]
    \centering
    \includegraphics[width=0.9\linewidth]{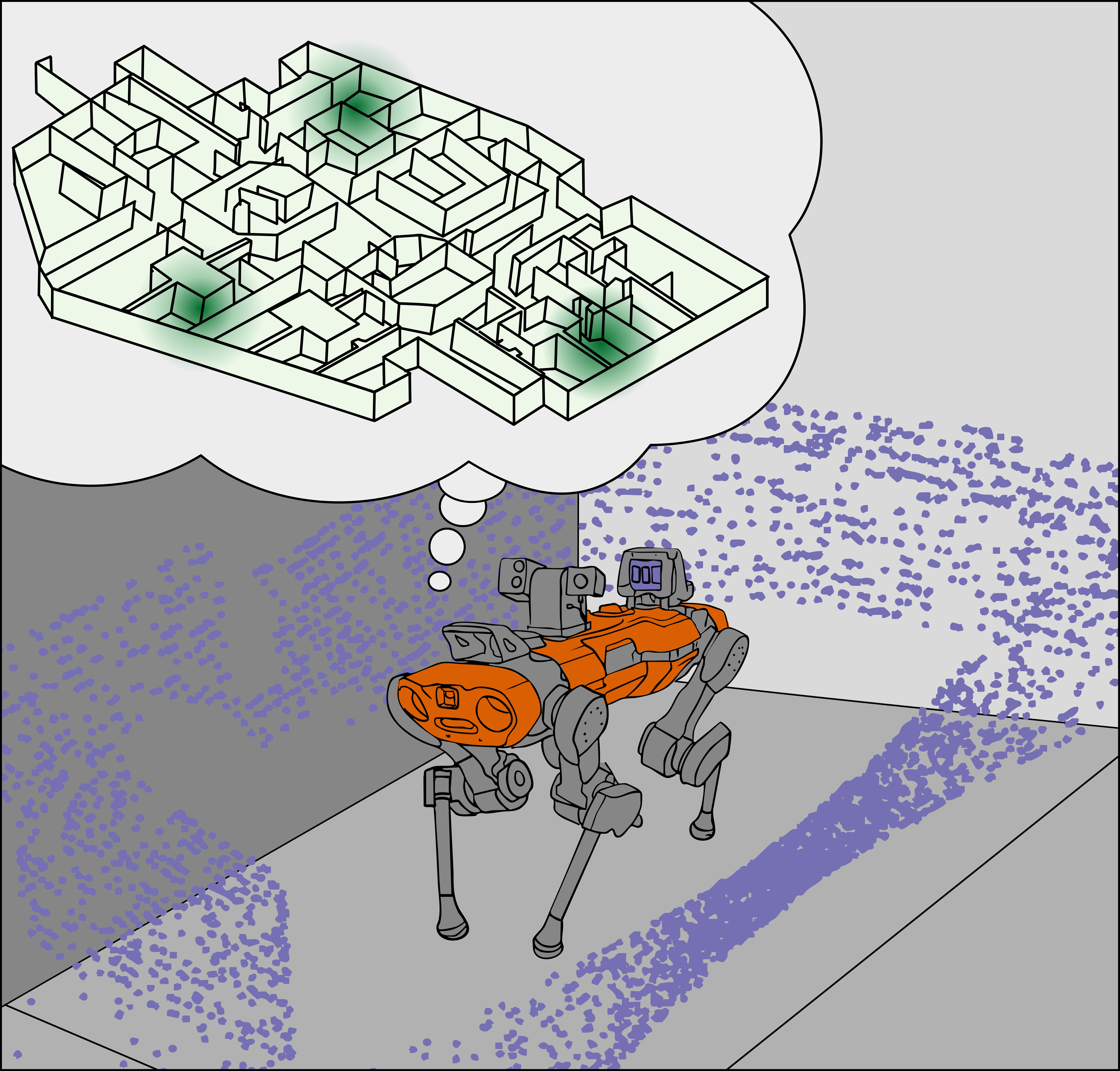}
    \caption{The robot’s LiDAR sensor captures a point cloud of the environment (purple points). Our trained model directly predicts the most likely positions in the building mesh that could have generated this observation. These candidate locations are visualized in green within the mesh. Figure created by Anna Bürgisser}
    \label{fig:motivation}
\end{figure}

Construction sites are difficult environments for localization due to a variety of factors; GPS is often unavailable indoors, ambiguity is introduced from self-similar locations, and they are often devoid of objects typically present in common indoor settings (e.g., furniture, decorations) that provide geometric and visual features that aid localization. LiDAR-SLAM methods can perform well but usually need a good initial pose and may fail in perceptually challenging areas such as staircases or corridors. The dynamic, rapidly changing nature of construction sites adds another degree of complexity since the pre-built map and current observations do not match.

In the presence of perceptually and geometrically similar environments, such as apartments with identical layouts, the localization system will have a hard time disambiguating the correct location. This is particularly challenging when using a single observation of the environment. We propose to address location self-similarity and extreme perceptual degradation of construction sites through temporal tracking of multiple hypotheses until such ambiguity can be resolved.

Particle filters~\cite{dellaert1999monte} allow tracking of multiple location hypotheses by using a set of particles to represent the potential robot state. Particle filters approximate the location distribution with a set of weighted particles that are propagated through motion models and updated with sensor observations. This approach, commonly known as \acf{mcl}, provides temporal tracking of multiple pose hypotheses and is widely used for achieving robust and precise global localization.\par

However, the out-of-the-box \ac{mcl} formulation cannot cope directly with the perceptual challenges of construction sites. Hence, in this work, we investigate how to improve the observation model of the particle filter to be able to robustly deal with a construction environment as well as how such an observation model can be incorporated into a particle filter framework.
This paper focuses on the localization of robots equipped with LiDAR sensors within laser-scanned meshes of construction environments, as illustrated in Figure \ref{fig:motivation}. We propose a system that combines a trained transformer-decoder-based place recognition model---which estimates the most likely robot location(s) from a single LiDAR scan---with a particle filter that incorporates both current observations and temporal state information to refine position estimates over time. Our framework is trained exclusively on synthetic LiDAR data, yet we demonstrate that it generalizes to real-world scans from the same environment.\par

The main contribution of this paper is a \acf{mcl} system that: (i) employs a trained place recognition model capable of predicting multiple location hypotheses from a single LiDAR scan; (ii) integrates scan-based uncertainty estimates into the observation model of the \ac{mcl}; (iii) introduces a resampling strategy that enables the particle filter to recover when the system becomes overconfident in an incorrect location, thereby enhancing robustness; and (iv) provides a detailed experimental evaluation on four real-world datasets.

%%%%%%%%%%%%%%%%%%%%%%%%%%%%%%%%%%%%%%%%%%%%%%%%%%%%%%%%%%%%%%%%%%%%%%%%%%%%%%%%
\section{Related Work}
\label{sec:related}

Global localization in 3D environments has been approached through a variety of point cloud embedding methods, ranging from hand-crafted representations to deep learning-based models. A widely adopted hand-crafted technique is ScanContext++~\cite{gkim-2019-ral}, which abstracts a LiDAR scan into polar-coordinate bins and records the maximum height within each bin. This compact descriptor enables efficient place recognition by comparing similarities across a database of previously observed scans, and it has become a common baseline for global localization tasks~\cite{krummenacher2025diffusion}.
MinkLoc3D~\cite{komorowski2021minkloc3d} leverages a learning-based, sparse voxelized encoding of the point cloud combined with sparse 3D convolutions to generate discriminative embeddings, allowing robust retrieval of similar places in large-scale databases. Another deep learning-based approach builds on PointNet++ \cite{qi2017pointnet++}, which hierarchically abstracts sets of points to capture local and global geometric structures. In PointLoc~\cite{wang2021pointloc}, this PointNet++ architecture is extended with self-attention modules, pooling operations, and a regression head to directly predict the global pose of the sensor. However, this approach does not incorporate hypothesis tracking and can produce only a single pose estimate per observation.

A different line of work is represented by OverlapNet~\cite{chen2022overlapnet}, which employs a Siamese neural network trained to estimate the degree of overlap between pairs of LiDAR scans. Instead of directly embedding scans into a feature space, the network predicts how much two observations share in common. This overlap estimation is then integrated into a \ac{mcl} framework, where the weight of each particle is computed by comparing the real LiDAR scan with a simulated scan generated from the particle’s pose hypothesis. This requires simulating a LiDAR scan for every particle at each update step, making the method computationally expensive and less suitable for real-time deployment, although this can partly be mitigated through hardware acceleration and parallelization.

Another approach directly predicting the robot’s pose from LiDAR point clouds was introduced by Krummenacher et al.~\cite{krummenacher2025diffusion}. Their method leverages a denoising diffusion probabilistic model, which iteratively refines an arbitrary initial position into a final location estimate. This iterative process can generate multiple distinct hypotheses for the robot's position. We adopt and extend their pipeline for synthesizing LiDAR data from a global mesh, and we use their method---referred to hereafter as the diffusion model---as our baseline for comparison. In their study, Krummenacher et al. further show that this diffusion-based approach outperforms the widely used ScanContext++ algorithm in localization tasks.

\ac{mcl} \cite{dellaert1999monte} is a well-established method that remains widely used for robot localization. It is an implementation of a particle filter to represent and update multiple hypotheses of the robot’s pose. The motion model propagates particles according to the robot’s movement, while the observation model assigns weights based on how well each particle’s pose explains the current measurement. The particle set is then resampled according to these weights to focus on the most plausible hypotheses.
Although effective, the original \ac{mcl} formulation is prone to mode collapse, also known as the particle depletion problem: once the particles converge to an incorrect hypothesis, the filter may become stuck in that state and fail to recover. In this work, we inject the location hypothesis reported by a learned place recognition model to recover from the particle depletion situation.

Only a limited number of works have focused on localization in construction site environments, particularly in the scan-to-map setting. Yin et al. \cite{yin2023semantic} propose using semantically augmented point clouds generated from \ac{bim} models as a reference, and then apply \ac{icp} registration to align incoming LiDAR scans. Blum et al. \cite{blum2021precise} address the sim-to-real gap by introducing outlier rejection and sensor fusion techniques to perform \ac{icp} on robust surface representations specifically tailored to construction sites. Dreher et al. \cite{dreher2021global} extract and summarize structural planes from LiDAR scans and match them against building meshes to achieve localization.
While these \acs{icp}-based approaches can be effective, they have a notable limitation: localization hypotheses are not tracked over time as only a single pose estimate is produced per observation. In contrast, our method combines a learned observation model with a particle filter, enabling efficient tracking of multiple hypotheses and scalable localization in complex construction environments.

%%%%%%%%%%%%%%%%%%%%%%%%%%%%%%%%%%%%%%%%%%%%%%%%%%%%%%%%%%%%%%%%%%%%%%%%%%%%%%%%
\section{Localization in Construction Site Meshes}
\label{sec:methods}

\begin{figure*}[ht]
    \centering
    \footnotesize
    \def\svgwidth{\textwidth}
    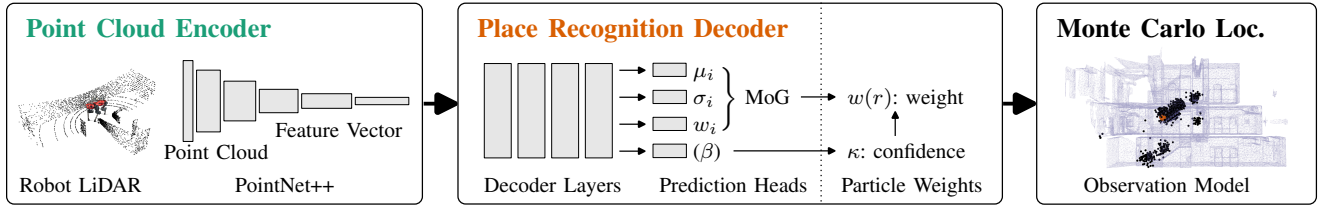
    \caption{Overview of the proposed localization pipeline. In the training stage, the robot is simulated inside the building mesh to generate synthetic LiDAR scans. Each scan is processed by a point cloud encoder and a place recognition model, which predicts the most likely robot locations from a single observation. In the deployment stage, the same encoder–decoder pipeline is applied, but the outputs are reinterpreted as an observation model: predictions are converted into particle weights that combine likelihoods with uncertainty estimates. These weights are then integrated into the \ac{mcl}, providing the observation model for localization.}
    \label{fig:pipeline}
\end{figure*}

The goal of our method is to build a localization pipeline that represents multiple pose hypotheses, tracks them over time, and avoids mode collapse. We adopt a \acl{mcl} framework with learned LiDAR-based observation models. In training, synthetic LiDAR scans are simulated inside the building mesh, embedded with a point cloud encoder, and mapped to likely robot locations by a place recognition model. At inference, this model is reinterpreted as the observation model for \acs{mcl}, providing particle weights that incorporate likelihoods and uncertainty, which are combined with the motion model, resampling, and prediction to achieve robust relocalization in the mesh environment. An overview of the complete pipeline is shown in Figure~\ref{fig:pipeline}. We discuss the details in the section below.

\subsection{Robot Simulation}
\label{sec:sim}

Our training pipeline begins by simulating the robot within the mesh environment, inspired by the work of Krummenacher et~al.~\cite{krummenacher2025diffusion}. We first determine valid poses for the robot inside the mesh and then, for each pose, simulate the LiDAR sensor’s point cloud by casting rays into the mesh from that position and orientation.

To accelerate training, we sample only poses the robot could realistically occupy in the mesh. Simple horizontal-plane detectors break down in complex areas such as staircases and ramps, so instead we detect standable surfaces directly from the mesh. We first extract the floors of the mesh, i.e., all positions where the quadrupedal robot could stand. To do this, we create a 2D grid covering the horizontal plane $(x,y)$ of the mesh, with a spacing of \SI{0.1}{\metre} between grid points. This grid is extruded vertically through the mesh, and all intersections (hit points) between the vertical lines and the mesh are recorded. We mark a hit point as a valid floor point if the next intersection above it lies between \SI{2}{\metre} and \SI{10}{\metre}, corresponding to typical ceiling heights. For each valid floor point, we translate it upward by a random value based on the LiDAR sensor height during deployment, and we apply a random orientation.

For each training pose, we generate a simulated LiDAR point cloud by ray tracing the sensor’s known scan pattern through the mesh and recording the intersection points. To increase realism, we add noise to both the ray directions and the measured distances. The resulting point cloud is then preprocessed by discarding points with ranges outside 0.5--20 m. Finally, the cloud is randomly downsampled to 4096 points to accelerate the following point cloud encoding process.

\subsection{Point Cloud Encoder}

The goal of this step is to extract a compact and informative representation from the point cloud. Following \cite{krummenacher2025diffusion}, we use a PointNet++ \cite{qi2017pointnet++} architecture as the feature extractor, which processes the point cloud through set abstraction layers that group points into progressively larger regions while learning local geometric features before aggregating them into a single global descriptor via max pooling. 
% Our implementation follows the standard PointNet++ design, with the exact configuration of the input, set abstraction, and grouping layers summarized in Table~\ref{tab:pointnet++}, which serves only to document the settings used in our pipeline.

% \begin{table}[ht]
%  \centering
%  \footnotesize
%  \caption{PointNet++ point cloud encoder layers}
%  \label{tab:pointnet++}
%  \begin{tabular}{lcccc}
%     \toprule
%     Layer Name & Points & Radius & Ratio & Features\\
%     \midrule
%     Input Point Cloud & 4096 & - & - & - \\
%     Set Abstraction 1 & 2048 & 0.5 & 0.5 & 128\\
%     Set Abstraction 2 & 512 & 1.0 & 0.25 & 256 \\
%     Set Abstraction 3 & 64 & 2.0 & 0.125 & 256\\
%     Group All & 1 & - & - & 256\\
%     \bottomrule
%  \end{tabular}
% \end{table}

% The following figure is placed here for layout reasons, but it is referenced and discussed in a later section.
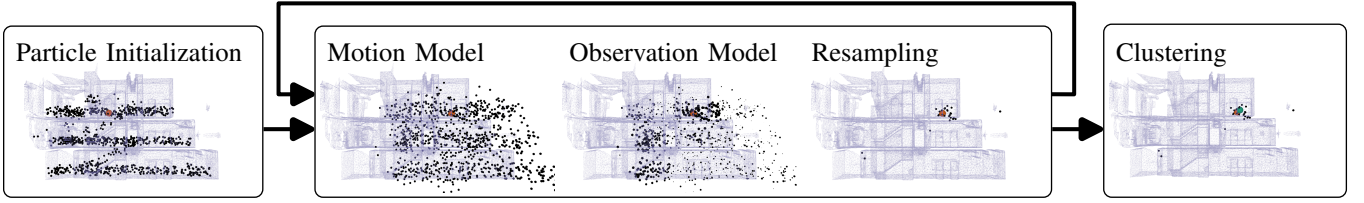
\begin{figure*}[ht]
    \centering
    \def\svgwidth{\textwidth}
    %% Creator: Inkscape 1.4.2 (ebf0e940d0, 2025-05-08), www.inkscape.org
%% Accompanies image file 'MCL.pdf' (pdf, eps, ps)
\begingroup%
  \makeatletter%
  \providecommand\color[2][]{%
    \errmessage{(Inkscape) Color is used for the text in Inkscape, but the package 'color.sty' is not loaded}%
    \renewcommand\color[2][]{}%
  }%
  \providecommand\transparent[1]{%
    \errmessage{(Inkscape) Transparency is used (non-zero) for the text in Inkscape, but the package 'transparent.sty' is not loaded}%
    \renewcommand\transparent[1]{}%
  }%
  \providecommand\rotatebox[2]{#2}%
  \newcommand*\fsize{\dimexpr\f@size pt\relax}%
  \newcommand*\lineheight[1]{\fontsize{\fsize}{#1\fsize}\selectfont}%
  \ifx\svgwidth\undefined%
    \setlength{\unitlength}{505.0826416bp}%
    \ifx\svgscale\undefined%
      \relax%
    \else%
      \setlength{\unitlength}{\unitlength * \real{\svgscale}}%
    \fi%
  \else%
    \setlength{\unitlength}{\svgwidth}%
  \fi%
  \global\let\svgwidth\undefined%
  \global\let\svgscale\undefined%
  \makeatother%
  \begin{picture}(1,0.14461411)%
    \lineheight{1}%
    \setlength\tabcolsep{0pt}%
    \put(0,0){\includegraphics[width=\unitlength,page=1]{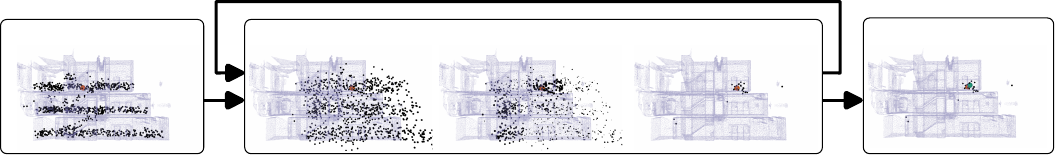}}%
    \put(0.00980038,0.10163224){\color[rgb]{0,0,0}\makebox(0,0)[lt]{\lineheight{1.25}\smash{\begin{tabular}[t]{l}Particle Initialization\end{tabular}}}}%
    \put(0.24144563,0.10163224){\color[rgb]{0,0,0}\makebox(0,0)[lt]{\lineheight{1.25}\smash{\begin{tabular}[t]{l}Motion Model\end{tabular}}}}%
    \put(0.42161415,0.10163224){\color[rgb]{0,0,0}\makebox(0,0)[lt]{\lineheight{1.25}\smash{\begin{tabular}[t]{l}Observation Model\end{tabular}}}}%
    \put(0.82828026,0.10163224){\color[rgb]{0,0,0}\makebox(0,0)[lt]{\lineheight{1.25}\smash{\begin{tabular}[t]{l}Clustering\end{tabular}}}}%
    \put(0.6017827,0.10163224){\color[rgb]{0,0,0}\makebox(0,0)[lt]{\lineheight{1.25}\smash{\begin{tabular}[t]{l}Resampling\end{tabular}}}}%
  \end{picture}%
\endgroup%

    \caption{\acl{mcl} pipeline. The construction site mesh is shown as a purple wireframe. Black points denote the particles of the filter, the orange point marks the ground-truth position, and the green dot indicates the cluster center. Particle initialization is performed only once at the beginning, whereas clustering is applied at every evaluation interval to obtain the current prediction.}
    \label{fig:mcl}
\end{figure*}

\subsection{Place Recognition Decoder}

The purpose of this step is to estimate the robot’s position from the point cloud features. To achieve this, we introduce a transformer-decoder-based architecture~\cite{vaswani2017attention}, which outputs a probability density function over the continuous 3D space rather than a single point estimate. This allows the model to naturally represent multiple hypotheses about the robot’s location. In our implementation, we fix the number of hypotheses to $K=5$, which corresponds to the five weighted Gaussian components ($\mu_i$, $\sigma_i$, $w_i$, where $i \in \{1, \dots, 5 \}$) used later in the predicted distribution.

The decoder operates on two inputs: the global point-cloud embedding of size $(1,256)$ produced by the encoder, and a set of $K=5$ trainable hypothesis queries ~\cite{jaegle2021perceiver} of size $(1,256)$. These queries are not tied to any specific sensor input but act as learnable biases that give the model room to represent several possible robot locations for the same scan. Intuitively, they provide "slots" for alternative explanations, which helps the system handle the ambiguities that often occur in construction environments. The decoder is composed of four transformer decoder layers, each with 256-dimensional features and four attention heads, and a dropout rate of 10\% is applied throughout to improve generalization.

In self-similar environments, a single LiDAR scan may correspond to several distinct locations---for example, in rooms with identical layouts or in long, featureless corridors. A unimodal estimate would force an early and often fragile decision, while a multi-modal distribution can represent all plausible locations until future observations resolve the ambiguity. We therefore model the robot’s position as a \ac{mog} distribution with $K=5$ components, each representing a distinct hypothesis:
\begin{equation}
p_{\text{mog}}(r \mid f) = \sum_{i=1}^K w_i \cdot \mathcal{N}(r \mid \mu_i, \sigma_i)
\end{equation}
The distribution parameters are obtained from linear prediction heads on the final decoder layer. The means are predicted as $\mu_i \in \mathbb{R}^3$. For the standard deviations, we predict one value for the $x$ and $y$ axes ($\sigma_i^x = \sigma_i^y$) and one for the $z$-axis, reducing the output to two positive values ($\sigma_i^x, \sigma_i^z \in \mathbb{R}^+$). The mixture weights $w_i \in [0,1]$ are predicted with a softmax layer, ensuring $\sum_{i=1}^K w_i=1$.

The model is trained using the negative log-likelihood (NLL) loss of the ground-truth position $r_{gt}$ under the predicted \acl{mog}:
\begin{equation}
    \mathcal{L}_{\text{NLL}} = -\log  \left( p_{\text{mog}}(r_{\text{gt}} \mid f) \right)
\end{equation}
This formulation is numerically stable, even when the likelihood of the ground-truth position is very small, since operating in log-space prevents underflow and allows the model to handle extreme probability values robustly.

\subsection{Place Recognition Decoder with Confidence Prediction}

In the challenging setting of construction environments, a single LiDAR scan may provide only weak evidence for the robot’s location. To ensure that predictions reflect this variability, we extend the previously introduced place recognition decoder with an additional prediction head that outputs a confidence parameter $\beta \in [0,1]$ via a sigmoid function. Intuitively, $\beta$ expresses how confident the model is in its \ac{mog} prediction: if $\beta \approx 1$, the \ac{mog} is trusted; if $\beta \approx 0$, the model falls back to a uniform likelihood.
The fallback likelihood represents what we would believe without any model---namely, that every valid position in the mesh is equally likely. If the navigable mesh volume is $V$, this uniform likelihood is $V^{-1}$. To approximate $V$, we voxelize the mesh at \SI{1}{\metre} resolution and count the occupied voxels.

Training encourages $\beta$ to choose between the \ac{mog} and uniform likelihoods, depending on which better explains the ground truth for the current embedding. In ambiguous, self-similar regions (e.g., staircases), the model can reduce $\beta$ and rely more on the fallback; in distinctive areas, it can increase $\beta$ and trust the \ac{mog}. The resulting objective is
\begin{equation}
\mathcal{L}_{\text{NLL conf.}} = - \log \Big(\beta \cdot p_{\text{mog}}(r_{\text{gt}} \mid f) + (1-\beta) \cdot V^{-1} \Big).
\end{equation}

\subsection{Monte Carlo Localization}

The goal of this step is to provide temporal consistency in the position estimates produced by the place recognition model. We adopt a particle filter framework, following the classical \ac{mcl} method proposed in \cite{dellaert1999monte}. A visual explanation of the \ac{mcl} pipeline is provided in Figure~\ref{fig:mcl}. In the following, we describe our adaptations to its key components: particle initialization, the motion model, the observation model, particle resampling, and the final pose estimation.

We represent the belief over the robot’s pose with $N = 1{,}000$ particles. Each particle encodes a full 6-DoF pose: the 3D position and a quaternion for orientation. To initialize the particles, we follow the same procedure used for generating training poses (see Section~\ref{sec:sim}): particles are placed at valid robot poses inside the mesh, ensuring that all initial states correspond to physically feasible positions.

A legged-inertial estimator can provide smooth odometry for the filter's motion model, and is ideal for construction sites since LiDAR- and vision-based odometry frequently diverges.
Since some datasets were collected by humans, kinematic odometry for these is not available.
Instead, to ensure a consistent motion model, we approximate motion by taking the ground-truth translation and rotation between two evaluation steps and add zero-mean Gaussian noise. The standard deviation of this noise is derived from the change in pose between the current and previous time steps together with the elapsed time, so that larger or faster motions and longer update intervals naturally result in higher uncertainty.

While the motion model captures how particles evolve over time, the observation model determines how well each particle explains the current sensor input, by assigning weight to the particles. We implement this directly with the trained place recognition decoder: its predicted \ac{mog} likelihood $p_{\text{mog}}(r \mid f)$ for position $r$ given embedding $f$ is combined with a uniform prior ($1/N$) to form particle weights,
\begin{equation}
w(r) = (1 - \kappa) \cdot \frac{1}{N} + \kappa \cdot p_{\text{mog}}(r \mid f).
\end{equation}

Here, $\kappa \in [0, 1]$ is a unified confidence that calibrates how much to trust the decoder output on a given scan. To compensate for the sim-to-real gap caused by training only on synthetic LiDAR data, we use a baseline value $\kappa_{\text{sim-to-real}} = 0.5$. The basic place recognition decoder fixes $\kappa = \kappa_{\text{sim-to-real}}$; the place recognition decoder with confidence prediction scales this baseline with the learned gate $\beta$: $\kappa = \kappa_{\text{sim-to-real}} \cdot \beta$. A lower $\kappa$ increases the influence of the uniform term, reducing weight variation across particles and making subsequent resampling less aggressive. This improves robustness in uncertain situations, as more hypotheses are retained for the next step.

We then perform systematic resampling~\cite{kuptametee2022review} on the normalized weights of the particles, a common resampling method in particle filters that helps retain some particles with low probabilities. In this method, all particle weights are laid out in a cumulative “stack.” A random starting point $s \in [0, 1/N]$ is chosen, and then $N$ samples are taken at fixed intervals of $1/N$ along the stack. 

Particle filters are prone to mode collapse, where the belief distribution concentrates on a single hypothesis. Once all particles support this hypothesis, the filter can only escape through the randomness of the motion model, which is often too weak to recover. To address this limitation, we modify the resampling step: after resampling, we replace the positions of $K = 5$ randomly chosen particles with the Gaussian means $\mu_i$ predicted by the place recognition model. In this way, the place recognition model’s own hypotheses are explicitly injected into the particle set, ensuring that alternative modes remain represented and enabling escape from mode collapse.

To evaluate performance and support downstream tasks, we require a single pose estimate from the particle filter, since the full set of $N=1{,}000$ hypotheses is not directly interpretable. To obtain this estimate, we apply the DBSCAN~\cite{ester1996density} clustering algorithm. We configure DBSCAN with a neighborhood radius of \SI{1}{\metre} and a minimum cluster size of 100 particles, which corresponds to one tenth of the total number of particles. When clusters are detected, we choose the cluster with the largest number of particles and take its centroid as the predicted position. If no valid clusters are found, we instead fall back to the mean position of all particles to provide a pose estimation.

%%%%%%%%%%%%%%%%%%%%%%%%%%%%%%%%%%%%%%%%%%%%%%%%%%%%%%%%%%%%%%%%%%%%%%%%%%%%%%%%
\section{Experimental Evaluation}
\label{sec:exp}

We design our experiments to demonstrate the capabilities of the proposed method and to validate its effectiveness in real-world construction scenarios. The results support three key claims:
(i) our \ac{mcl}-based approach consistently outperforms the baselines in construction environments,
(ii) confidence estimation improves robustness in the challenging conditions of construction sites, where self-similarity, clutter, and incomplete structures make localization difficult, and
(iii) the proposed resampling strategy enables recovery when the particle filter collapses into an incorrect mode.

%% Note 1: It MUST be always crystal clear (a) WHY an experiment is there
%% (e.g., to support a claim, to show that the approaches useful for real-word
%% systems, to show the performance, or to provide a baseline comparison), (b)
%% WHAT it wants to show (which claim/property exactly), and (c) HOW it aims at 
%% showing this. This is ESSENTIAL for a good evaluation. Think about when BEFORE
%% designing an experiment.  IMPORTANT: Every experiment MUST start with something 
%% like:  The next experiment is presented to show \dots and thus for supporting our 
%% first claim.

%% Note 2: Start with the most important/impressive experiment first. Make
%% his a key story of the paper. Keep the order of the claims, i.e., re-order
%% claims in the intro/before if needed. 

\subsection{Datasets}

\begin{figure}[ht]
    \centering
    \def\svgwidth{\linewidth}
    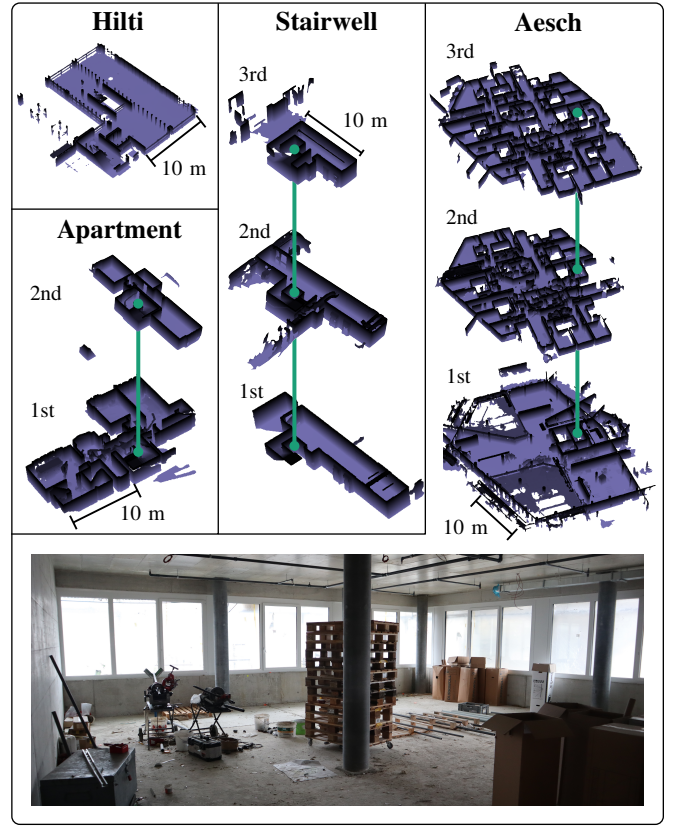
    \caption{Overview of the four datasets used for evaluation. The green line marks the staircase used for the multifloor datasets. A \SI{10}{\metre} scale bar provides spatial reference. For the Aesch dataset, a representative image of the environment is included.}
    \label{fig:datasets}
\end{figure}

% for better layout here
\begin{table*}[ht]
    \footnotesize
    \caption{Monte Carlo Localization (MCL) results. Recall indicates the proportion of predictions within the specified radius. Two evaluation modes are compared: \emph{Continuous}, where particles are initialized once and evaluated at every step, and \emph{Reinit}, where particles are reinitialized every 10 steps and metrics are reported at step 10.}
    \label{tab:mcl}
    \centering
    \vspace{-2mm}
     \begin{tabular}{ll|cc|cc|cc|cc|cc|}
 \toprule
 \multirow{2}{*}{Dataset} & \multirow{2}{*}{Model}
 & \multicolumn{2}{c|}{Recall@\SI{4}{\metre} $\uparrow$} & \multicolumn{2}{c|}{Recall@\SI{2}{\metre} $\uparrow$} & \multicolumn{2}{c|}{Recall@\SI{1}{\metre} $\uparrow$} & \multicolumn{2}{c|}{Recall@\SI{0.5}{\metre} $\uparrow$} & \multicolumn{2}{c|}{Error [\SI{}{\metre}] $\downarrow$} \\
 & & Cont. & Reinit & Cont. & Reinit & Cont. & Reinit & Cont. & Reinit & Cont. & Reinit \\
 \midrule

 \multirow{4}{*}{Hilti}
 & ScanContext++ & 0.57 & 0.69 & 0.29 & 0.54 & 0.29 & 0.38 & 0.00 & 0.15 & 5.56 & 5.92 \\
 & Diffusion & 0.88 & 0.78 & 0.75 & 0.33 & 0.43 & 0.22 & 0.18 & 0.11 & 2.46 & 3.02 \\
 & Transformer Decoder & 0.93 & \textbf{0.89} & 0.92 & 0.67 & 0.49 & 0.44 & 0.13 & 0.11 & 1.54 & 2.57 \\
 & \qquad w/ Confidence Pred. & \textbf{0.99} & \textbf{0.89} & \textbf{0.99} & \textbf{0.89} & \textbf{0.91} & \textbf{0.89} & \textbf{0.56} & \textbf{0.56} & \textbf{0.64} & \textbf{1.48}  \\
 \midrule

 \multirow{4}{*}{Apartment}
 & ScanContext++ & 0.80 & 0.52 & 0.30 & 0.24 & 0.00 & 0.00 & 0.00 & 0.00 & 3.29 & 4.07 \\
 & Diffusion & 0.95 & 0.93 & 0.86 & 0.76 & 0.69 & 0.69 & 0.34 & 0.28 & 1.31 & 1.20 \\
 & Transformer Decoder & 0.96 & 0.93 & 0.85 & 0.79 & 0.81 & 0.72 & 0.57 & 0.52 & 0.96 & 1.31 \\
 & \qquad w/ Confidence Pred. & \textbf{0.99} & \textbf{1.00} & \textbf{0.91} & \textbf{0.90} & \textbf{0.86} & \textbf{0.85} & \textbf{0.78} & \textbf{0.75} & \textbf{0.57} & \textbf{0.90} \\ 
 \midrule

 \multirow{4}{*}{Stairwell}
 & ScanContext++ & 0.63 & 0.30 & 0.13 & 0.04 & 0.13 & 0.04 & 0.00 & 0.00 & 3.97 & 4.63 \\
 & Diffusion & 0.70 & 0.65 & 0.62 & 0.61 & 0.51 & 0.52 & 0.20 & 0.17 & 2.62 & 2.82 \\
 & Transformer Decoder & 0.85 & \textbf{0.78} & 0.67 & 0.60 & 0.61 & 0.57 & 0.40 & 0.39 & 1.78 & 2.07 \\
 & \qquad w/ Confidence Pred. & \textbf{0.93} & 0.74 & \textbf{0.83} & \textbf{0.65} & \textbf{0.73} & \textbf{0.61} & \textbf{0.48} & \textbf{0.52} & \textbf{1.25} & \textbf{1.62} \\ 
 \midrule
 
 \multirow{4}{*}{Aesch}
 & ScanContext++ & 0.04 & 0.03 & 0.00 & 0.00 & 0.00 & 0.00 & 0.00 & 0.00 & 9.04 & 10.13 \\
 & Diffusion & 0.58 & 0.49 & 0.34 & 0.32 & 0.14 & 0.15 & 0.03 & 0.03 & 5.66 & 6.57 \\
 & Transformer Decoder & 0.68 & \textbf{0.66} & \textbf{0.53} & 0.46 & 0.33 & 0.24 & 0.13 & 0.07 & \textbf{4.00} & \textbf{5.76} \\
 & \qquad w/ Confidence Pred. & \textbf{0.70} & 0.62 & \textbf{0.53} & \textbf{0.47} & \textbf{0.42} & \textbf{0.32} & \textbf{0.14} & \textbf{0.15} & 4.44 & 5.96 \\ 
 \bottomrule

 \end{tabular}
\end{table*}

To demonstrate robustness across typical construction scenarios, we evaluate on four real-world datasets---three from active sites---spanning different scales, floor counts, and acquisition setups. Together they cover small to large indoor volumes (\SI{1184}{\cubic\metre}-\SI{9719}{\cubic\metre}), single to three floors, and both robot and handheld trajectories. Figure~\ref{fig:datasets} shows the floor layouts of all datasets. In our work, we use reality-captured meshes obtained directly on-site with a Leica BLK2GO laser scanner~\cite{dlesk2022comparison}, as often done in modern construction projects~\cite{trzeciak2022conslam}, and employ these meshes as maps in our localization pipeline. Ground truth trajectories were estimated using Open3D SLAM~\cite{jelavic2022open3d}, localizing scans against preexisting environment maps. An accurate initial pose is found for each, and the algorithm is tuned offline for high accuracy; thus, it is not a viable online method.

\textit{Hilti Dataset.} The exp06 sequence from the Hilti 2022 challenge~\cite{hilti} captures a large, mostly open indoor area with few partitions. LiDAR was recorded with a Hesai PandarXT-32; the global map was built using a Z+F Imager 5016. It has a relatively smaller volume of \SI{1193}{\cubic\metre} and serves as a baseline with low self-similarity but limited structural cues.

\textit{Apartment Dataset.} The indoor sequence of the CON-3 mission from GrandTour~\cite{frey2025boxi} covers two floors of an active apartment construction. The trajectory was recorded with an ANYmal quadrupedal robot, and the map was captured using a handheld Leica BLK2GO~\cite{dlesk2022comparison}. Its volume of \SI{1184}{\cubic\metre} is also small, but the presence of smaller rooms and higher structural complexity across multiple floors distinguishes this scenario from the Hilti dataset.

\textit{Stairwell Dataset.} This is the only dataset not recorded on a construction site. The environment spans three floors connected by a staircase around an elevator, and we include it because of its strong self-similarities and bare concrete walls without furniture. Collected with the same instruments as the Apartment dataset, it has a volume of \SI{2470}{\cubic\metre}.

\textit{Aesch Dataset.} Introduced by~\cite{krummenacher2025diffusion}, this active construction site spans three floors, two of which have near-identical room layouts. The map was captured like in the Apartment dataset, while the trajectory was recorded with a handheld sensor box with a Hesai PandarXT-32. At \SI{9719}{\cubic\metre}, it is the largest and most self-similar setting, providing a stringent test of scalability and robustness under strong ambiguity.

\subsection{Baseline}

We compare our approach against two baselines: a ScanContext++~\cite{gkim-2019-ral} place recognition method and the diffusion model of~\cite{krummenacher2025diffusion}, as well as a random predictor for reference.

For the ScanContext++ baseline, we simulate 1000 robot poses and LiDAR scans as described in Section~\ref{sec:sim} to build a descriptor database. At inference, descriptors are matched to the database. For the place recognition evaluation, only the single best-matching pose is used. For the \ac{mcl} evaluation, the top $K=5$ matches are retained and reformulated as a \acf{mog}, where the selected poses define the means, the standard deviations are fixed to $\sigma_i=0.5$, and the mixture weights are uniform ($w_i = 1/K$). The confidence parameter is set to $\kappa_{\text{sim-to-real}} = 0.5$.

For the diffusion baseline, we replace our decoder-based place recognition module with the model of~\cite{krummenacher2025diffusion}, keeping all other components unchanged (Section~\ref{sec:methods}). The diffusion model generates a single position hypothesis per inference pass. However, since the denoising process starts from a random initialization, multiple distinct hypotheses can be obtained by running inference several times on the same point cloud embedding. To ensure a fair comparison with our decoder-based approach, we also generate $K=5$ hypotheses and convert them into a \acf{mog} using the same parameters as above for fairness.

Finally, the random baseline samples poses uniformly from the simulated training set (Section~\ref{sec:sim}).

\subsection{Implementation Details}
All model weights are randomly initialized. We use the Adam optimizer with a learning rate of $1 \times 10^{-4}$ and a batch size of 16. Each training epoch consists of 512 steps. Depending on the dataset size, the models are trained for different numbers of epochs: Hilti for 500 epochs (10.8 h), Apartment for 500 epochs (15.5 h), Stairwell for 400 epochs (10.3 h), and Aesch for 1,000 epochs (31.7 h). All training was conducted on a single NVIDIA TITAN RTX GPU. The plain decoder model without extensions contains 2.9M trainable parameters, corresponding to an estimated model size of 11.5 MB.  

% For the third claim:
\begin{figure*}[ht]
    \centering
    \begin{tabular}{ccc}
        \includegraphics[width=0.30\textwidth]{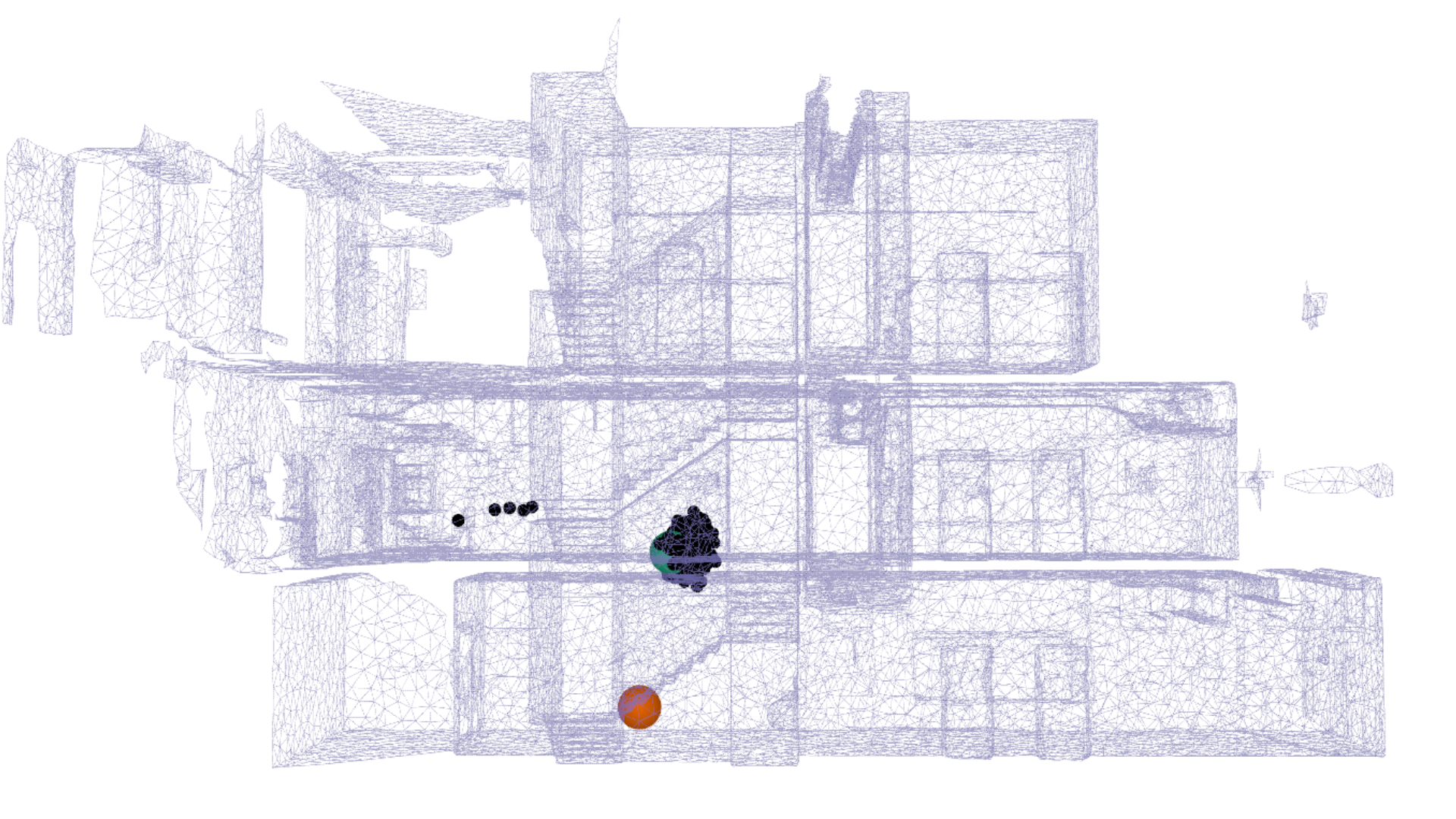} &
        \includegraphics[width=0.30\textwidth]{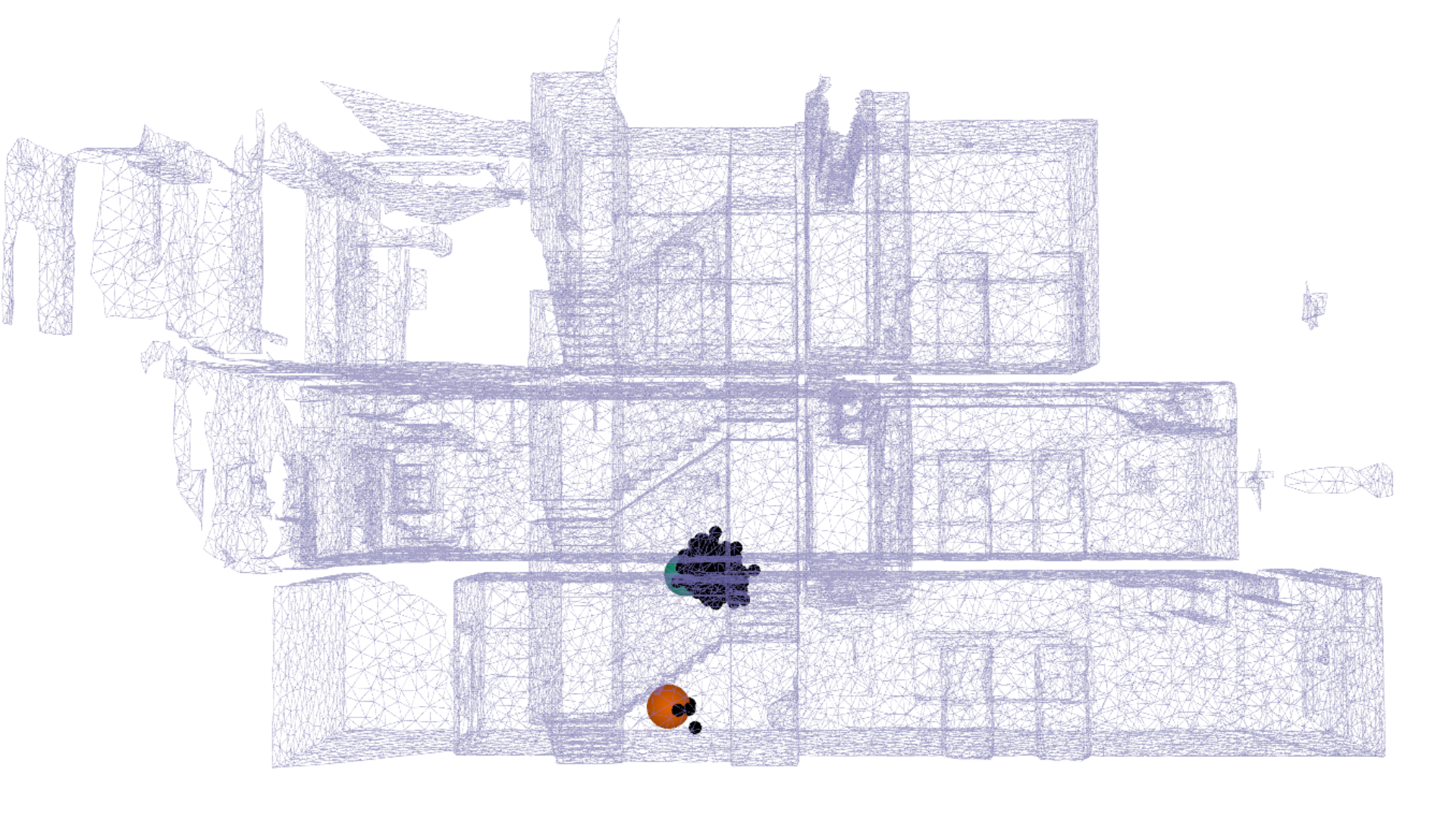} &
        \includegraphics[width=0.30\textwidth]{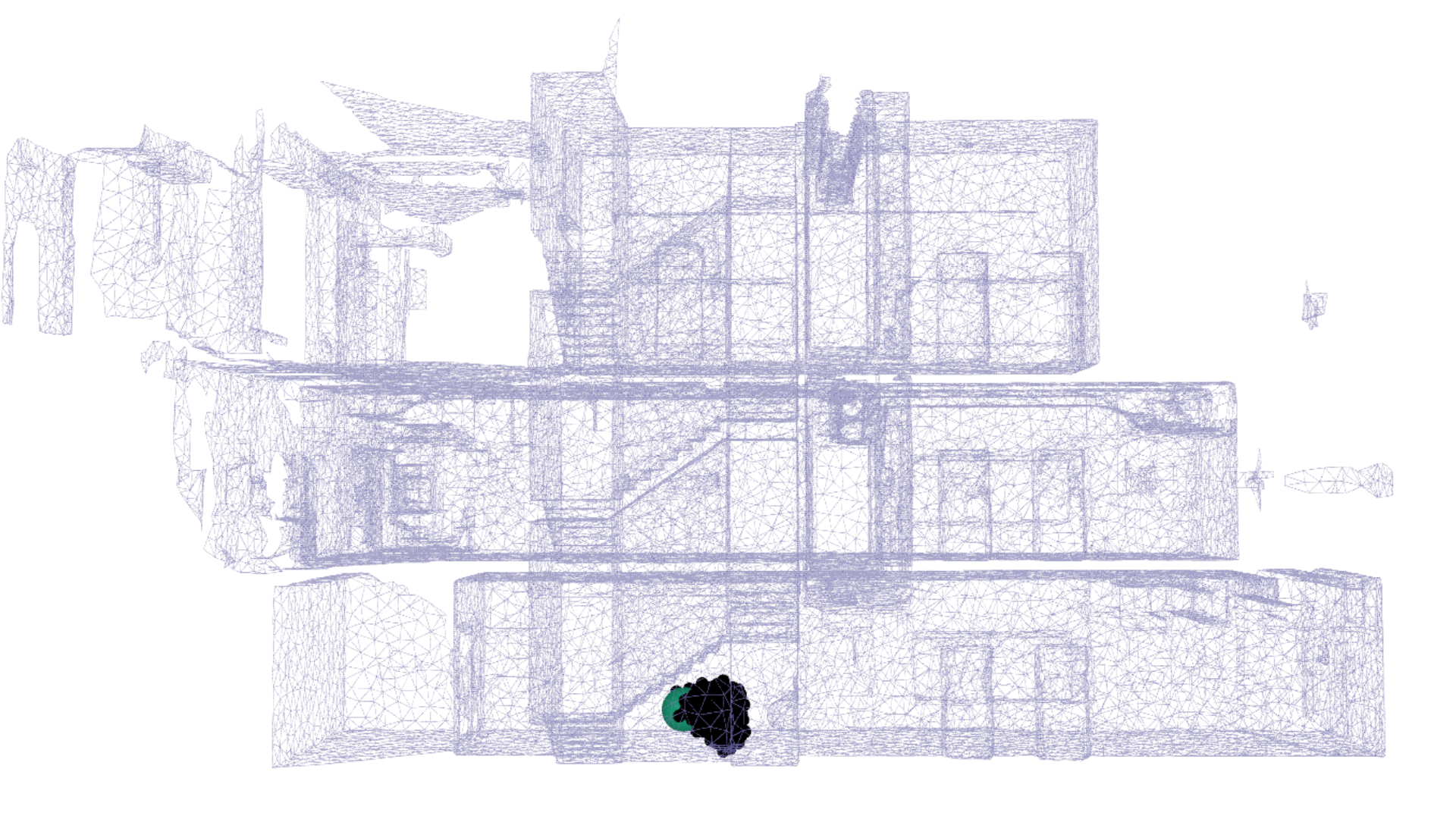} \\
    \end{tabular}
    \caption{Example of the model recovering from mode collapse. Black points show the particles of the filter and the orange point marks the ground-truth position within the building mesh. From left to right: illustration of mode collapse, reseeding of the particles in resampling stage, further \ac{mcl} steps.}
    \label{fig:mode_collapse}
\end{figure*}

\begin{table*}[ht]
    \centering
    \caption{Place recognition results. Recall measures the proportion of predictions falling within the specified radius, evaluated with respect to the maximum-likelihood position.}
    \label{tab:predictions}
    \footnotesize
    \vspace{-2mm}
         \begin{tabular}{llccccc}
     \toprule
     Dataset & Model & Recall@\SI{4}{\metre} $\uparrow$ & Recall@\SI{2}{\metre} $\uparrow$ & Recall@\SI{1}{\metre} $\uparrow$ & Recall@\SI{0.5}{\metre} $\uparrow$  & Error [\SI{}{\metre}]  $\downarrow$ \\ \hline
    
     \multirow{5}{*}{Hilti}
     & Random & 0.13 & 0.03 & 0.00 & 0.00 &  9.19\\
     & ScanContext++ & 0.39 & 0.24 & 0.12 & 0.02 & 7.49 \\
     & Diffusion & \textbf{0.70} & 0.49 & 0.23 & 0.05  & \textbf{4.15}\\
     & Transformer Decoder & 0.58 & 0.43 & 0.25 & 0.07 & 5.00 \\
     & \qquad w/ Confidence Pred. & 0.68 & \textbf{0.60} & \textbf{0.47} & \textbf{0.20} & 4.53 \\
     \midrule
    
     \multirow{5}{*}{Apartment}
     & Random & 0.22 & 0.06 & 0.01 & 0.00 & 6.68 \\
     & ScanContext++ & 0.24 & 0.10 & 0.07 & 0.02 & 6.33 \\
     & Diffusion & \textbf{0.93} & 0.74 & 0.50 & 0.18 & 1.73 \\
     & Transformer Decoder & 0.89 & 0.72 & 0.60 & 0.39 & 1.76\\
     & \qquad w/ Confidence Pred. & 0.91 & \textbf{0.79} & \textbf{0.71} & \textbf{0.50} & \textbf{1.53}\\ 
     \midrule
    
     \multirow{5}{*}{Stairwell}
     & Random & 0.15 & 0.04 & 0.01 & 0.00 & 7.33\\
     & ScanContext++ & 0.52 & 0.39 & 0.29 & 0.14 & 4.15 \\
     & Diffusion & 0.64 & 0.55 & 0.38 & 0.17 & 3.18\\
     & Transformer Decoder & 0.72 & 0.60 & 0.52 & 0.34 & 2.66 \\
     & \qquad w/ Confidence Pred. & \textbf{0.81} & \textbf{0.63} & \textbf{0.57} & \textbf{0.42} & \textbf{2.07} \\ 
     \midrule
     
     \multirow{5}{*}{Aesch} 
     & Random & 0.02 & 0.01 & 0.00 & 0.00 & 15.53 \\
     & ScanContext++ & 0.08 & 0.03 & 0.01 & 0.00 & 13.33 \\
     & Diffusion & 0.51 & 0.27 & 0.10 & 0.02 & \textbf{6.10}\\
     & Transformer Decoder & \textbf{0.55} & 0.33 & 0.15 & 0.03 & 6.48 \\
     & \qquad w/ Confidence Pred. & 0.53 & \textbf{0.37} & \textbf{0.19} & \textbf{0.05} & 6.60 \\ 
     \bottomrule
    
     \end{tabular}
\end{table*}

\subsection{Monte Carlo Localization Evaluation}

This section evaluates the full pipeline---combined place recognition model with \ac{mcl}---using the final pose obtained by clustering particles using DBSCAN. Table~\ref{tab:mcl} reports mean localization error and recall at several distance thresholds for two evaluation modes: \emph{Continuous}, where particles are initialized once at the start and propagated with predictions at roughly 1 Hz; and \emph{Reinit}, where particles are reinitialized every 10 steps (about 10 s) and performance is measured at step 10. The \emph{Reinit} setting forces regular restarts, exposing the particle filter to diverse initial conditions.

% For the first claim:
We first compare the transformer decoder with confidence prediction (with $\beta$) against the diffusion and ScanContext++ baselines in the \emph{Continuous} setting. Our model achieves a recall at \SI{1}{\metre} of 91\% on Hilti and 86\% on Apartment, outperforming the diffusion baseline (43\%, 69\%) and the ScanContext++ baseline (29\%, 0\%). On the more self-similar datasets, Stairwell and Aesch, \SI{1}{\metre} recall reaches 73\% and 42\% versus 51\% and 14\% for diffusion and 13\% and 0\% for the ScanContext++. The same trend also appears in the \emph{Reinit} setting, indicating that the method does not depend on favorable initial conditions. Overall, the results show improved \SI{1}{\metre} recall for the transformer decoder with confidence prediction across all datasets compared to the baselines, from less ambiguous to highly self-similar settings.

% For the second claim:
We next compare the transformer decoder with confidence prediction to the same model without confidence prediction in the \emph{Continuous} setting. The variant without confidence prediction achieves \SI{1}{\metre} recall of 49\% on Hilti, 81\% on Apartment, 61\% on Stairwell, and 33\% on Aesch. Although these results remain above the diffusion and ScanContext++ baselines, they fall short of the confidence-augmented model by 42, 5, 12, and 9 percentage points, respectively. Similar relative differences are observed in the \emph{Reinit} mode, confirming that incorporating data-driven confidence prediction consistently improves performance of the \ac{mcl} pipeline.

Fig.~\ref{fig:mode_collapse} supports our third claim---recovery of the filter from mode collapse. In the left image, particles have concentrated around an incorrect position. In subsequent steps (center, right), we inject hypotheses from the place recognition output: a fraction of particles is re-seeded at the means of the predicted \ac{mog} after the resampling step. This targeted diversification repopulates alternative modes and shifts probability mass toward the correct region as new observations arrive. By comparison, standard \ac{mcl}~\cite{dellaert1999monte} relies solely on motion model noise to escape incorrect mode collapse.

\subsection{Place Recognition Evaluation}

In this section, we compare the place recognition performance of the different models. For models that output multiple pose hypotheses (our method and the diffusion baseline), the predicted \acp{mog} are reduced to a maximum a posteriori (MAP) estimate. Since no closed-form solution exists, we approximate the MAP by evaluating the likelihood at the Gaussian means and selecting the mean with the highest value. Since the ScanContext++ baseline directly retrieves candidate matches, we select the position with the highest similarity score. This yields a single position estimate for each method and ensures a consistent comparison. We note that our method achieves an order-of-magnitude faster inference time than the diffusion baseline (0.018\,s vs.\ 0.11\,s per call), supporting its suitability for real-time deployment.

As shown in Table~\ref{tab:predictions}, the transformer decoder with confidence prediction achieves the highest recall at 2 m, 1 m, and 0.5 m. Notably, in this place recognition evaluation, the predicted confidence does not affect the MAP estimate. Therefore, the improvement over the decoder without confidence prediction likely stems from the confidence-augmented loss function.

\begin{figure}[ht]
    \centering
    \def\svgwidth{\linewidth}
    %% Creator: Inkscape 1.4.2 (ebf0e940d0, 2025-05-08), www.inkscape.org
%% Accompanies image file 'Confidence.pdf' (pdf, eps, ps)
\begingroup%
  \makeatletter%
  \providecommand\color[2][]{%
    \errmessage{(Inkscape) Color is used for the text in Inkscape, but the package 'color.sty' is not loaded}%
    \renewcommand\color[2][]{}%
  }%
  \providecommand\transparent[1]{%
    \errmessage{(Inkscape) Transparency is used (non-zero) for the text in Inkscape, but the package 'transparent.sty' is not loaded}%
    \renewcommand\transparent[1]{}%
  }%
  \providecommand\rotatebox[2]{#2}%
  \newcommand*\fsize{\dimexpr\f@size pt\relax}%
  \newcommand*\lineheight[1]{\fontsize{\fsize}{#1\fsize}\selectfont}%
  \ifx\svgwidth\undefined%
    \setlength{\unitlength}{246bp}%
    \ifx\svgscale\undefined%
      \relax%
    \else%
      \setlength{\unitlength}{\unitlength * \real{\svgscale}}%
    \fi%
  \else%
    \setlength{\unitlength}{\svgwidth}%
  \fi%
  \global\let\svgwidth\undefined%
  \global\let\svgscale\undefined%
  \makeatother%
  \begin{picture}(1,0.2863821)%
    \lineheight{1}%
    \setlength\tabcolsep{0pt}%
    \put(0,0){\includegraphics[width=\unitlength,page=1]{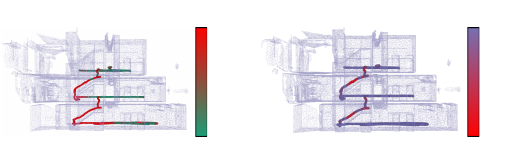}}%
    \put(0.02113821,0.24410568){\color[rgb]{0,0,0}\makebox(0,0)[lt]{\lineheight{1.25}\smash{\begin{tabular}[t]{l}\textbf{Place Recognition Error}\end{tabular}}}}%
    \put(0.53902437,0.24410568){\color[rgb]{0,0,0}\makebox(0,0)[lt]{\lineheight{1.25}\smash{\begin{tabular}[t]{l}\textbf{Confidence ($\beta$)}\end{tabular}}}}%
    \put(0.42248779,0.20223576){\color[rgb]{0,0,0}\makebox(0,0)[lt]{\lineheight{1.25}\smash{\begin{tabular}[t]{l}4 m\end{tabular}}}}%
    \put(0.42276421,0.02215447){\color[rgb]{0,0,0}\makebox(0,0)[lt]{\lineheight{1.25}\smash{\begin{tabular}[t]{l}0 m\end{tabular}}}}%
    \put(0.95121947,0.02288612){\color[rgb]{0,0,0}\makebox(0,0)[lt]{\lineheight{1.25}\smash{\begin{tabular}[t]{l}0\end{tabular}}}}%
    \put(0.95121947,0.20223576){\color[rgb]{0,0,0}\makebox(0,0)[lt]{\lineheight{1.25}\smash{\begin{tabular}[t]{l}1\end{tabular}}}}%
    \put(0,0){\includegraphics[width=\unitlength,page=2]{Confidence.pdf}}%
  \end{picture}%
\endgroup%

    \caption{Confidence visualization. Left: Place prediction error, color-coded from \SI{0}{\metre} to \SI{4}{\metre}. Right: predicted values of the confidence term $\beta$, ranging from 0 (low confidence) to 1 (high confidence).}
    \label{fig:confidence}
\end{figure}

To assess the transformer decoder with confidence prediction, we compare its predicted $\beta$ values with the localization error of the map estimate in Fig.~\ref{fig:confidence}. The model assigns lower $\beta$ scores in ambiguous regions, such as hallways and staircases, which coincide with higher map errors. This alignment indicates that the confidence prediction captures areas of structural ambiguities in the meshes.

%%%%%%%%%%%%%%%%%%%%%%%%%%%%%%%%%%%%%%%%%%%%%%%%%%%%%%%%%%%%%%%%%%%%%%%%%%%%%%%%
\section{Conclusion}
\label{sec:conclusion}

% In this paper, we presented a novel approach to\dots
% Our approach operates \dots  Our method exploits \dots
% This allows us to successfully \dots
% We implemented and evaluated our approach on different datasets
% and provided comparisons to other existing techniques and supported
% all claims made in this paper. The experiments suggest that \dots

In this work, we introduce a global localization pipeline that combines a transformer decoder-based place recognition model with an adapted \ac{mcl} framework. The system is designed for the challenging conditions of construction sites, offering robustness in self-similar environments and the ability to recover from mode collapse. It requires only the building mesh and LiDAR characteristics, eliminating the need for real-world training data. Our implementation demonstrates consistently superior performance compared to both diffusion-based and ScanContext++ baselines.

Despite these successes, our work has several limitations. First, the system depends on reality-captured meshes that were manually acquired on-site, rather than on \ac{bim} models. While this choice provides an accurate reference and simplifies the localization task, it requires an additional data collection step before deployment. Second, our method has not yet been thoroughly evaluated in rapidly changing environments, where large discrepancies may arise between the reference mesh and the robot’s LiDAR scans.

\bibliographystyle{plain_abbrv}

\bibliography{references}
% \bibliography{references_unabbreviated}

\end{document}